# Constructionist Steps Towards an Autonomously Empathetic System

**Trevor Buteau**
Fordham University
New York, NY 10023
tbuteau@fordham.edu

**Damian Lyons**
Fordham University
New York, NY 10023
dlyons@fordham.edu

## Abstract
Prior efforts to create an autonomous computer system capable of predicting what a human being is thinking or feeling from facial expression data have been largely based on outdated, inaccurate models of how emotions work that rely on many scientifically questionable assumptions. In our research, we are creating an empathetic system that incorporates the latest provable scientific understanding of emotions: that they are constructs of the human mind, rather than universal expressions of distinct internal states. Thus, our system uses a user-dependent method of analysis and relies heavily on contextual information to make predictions about what subjects are experiencing. Our system's accuracy and therefore usefulness are built on provable ground truths that prohibit the drawing of inaccurate conclusions that other systems could too easily make.

## ACM Classification Keywords
• **Human-centered computing~HCI theory, concepts and models**   • Human-centered computing~User centered design   • **Applied computing~Psychology**   • Computing methodologies~Activity recognition and understanding

**Constructed Emotion, Example**

Your throat is tight, your heart is racing, tears are running down your cheeks – so you access a series of emotional concepts (why are you feeling this way?) based on what you know about *the context in which these feelings are occurring*. At the wedding ceremony of a loved one – what you've been taught is a "happy" occasion – you will likely predict that these internal sensations mean you're feeling "happiness." On the other hand, at a funeral – what you've learned is a "sad" occasion – you will likely predict based on the *exact same internal sensations* that you are feeling "sadness."

While "the classical theory [says] we have [many] emotion circuits in our brains, and each [causes] a distinct set of changes, that is, a fingerprint." [4], Barrett's research (and the above example) show this simply isn't how emotions work.

**Introduction**

There is much existing research in the fields of affective computing and psychology which aims to predict what people are thinking and feeling from their facial expressions and other physiological data [10,11,17]. Much of it is based on the highly problematic "Classical Theory of Emotions", in which emotions are believed to be essential, discrete reactions of our bodies to changes in our environment [4], and therefore should be detectable in individuals by a machine once it learns what that emotion looks like for a general population. The research of Ekman and Friesen [6] is often held up as proof that this is how emotions work, and that, at least for the "basic" emotions that Ekman and Friesen "identified" (happiness, anger, sadness, surprise, fear, and disgust), they should be recognizable by anyone.

Thus, typically, affective computing relies on supervised learning systems using large databases of multi-subject data. The initially accurate results of this approach seem to support the Classical Theory of Emotions, but inter-database testing (testing a system trained on one database on a different database) can result in significant decreases in accuracy [8, 9, 12, 14] (particularly for "spontaneous" or "non-posed" subject data), suggesting that the assumptions of universality made by the Classical Theory of Emotion are wrong.

Our work is based on the newer "Theory of Constructed Emotions," (TCE) [4] which is a more scientifically provable and consistent framework for understanding human emotion and affect. According to neuroscientist Dr. Lisa Feldman Barrett, originator and lead proponent of the theory, *emotions are mental concepts we learn at a young age which help us predict the meaning of things in our environment, not automatic, pre-wired response systems in our brains that react the same way for all people. Barrett says the only way we can accurately predict what another person is feeling from looking at their face or other physiological signals is by also being given some contextual information*: what is happening to that person when they make that face? (see Constructed Emotion, Example sidebar) [4].

This understanding of how emotions are constructed in the brain has profound implications for how they can be predicted by a computer: since there is no behavioral, physiological, or neurological "universal fingerprint" for even a single emotion, emotions are not something that can be "detected" or "measured" in humans (no matter how we track facial expressions, changes in heart rate, changes in neuronal firing, etc.) [4]. Thus, any supervised learning system that compares the expressions of one individual to a population of others will be limited in its accuracy and usefulness.

We hold, however, that the "prediction" of a subject's internal experience by a computer is possible given that the computer system follows essentially the same mechanism that we use to predict our own emotions as individuals: we sense a series of biological signals and cross-check that combination of signals with what we know those signals have indicated in the past, and what we know about the context in which we are feeling them now.

In this paper, we describe our current system – which has already been tested with early results on an existing dataset of human behavior [16] – as well as a future use case/complete implementation of our system: predicting player choices in a game of Texas Hold 'Em (Poker) based on their behavior.

**Subject Responses, "T10" Task, MMSE-HR**

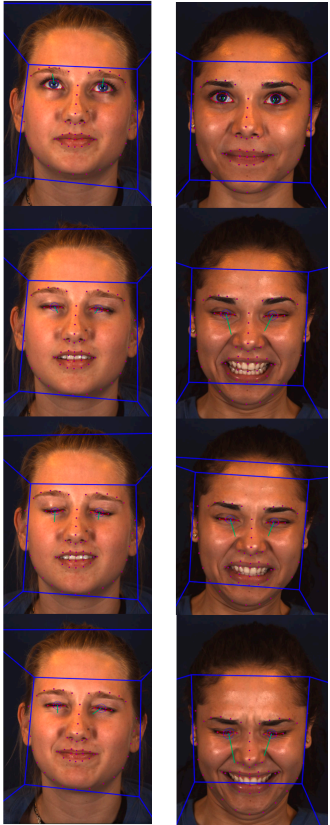

Figure 1: Images of two subjects' faces (processed by OpenFace) before and during the T10 protocol show great variance between subjects, but much consistency from one dart stimulus to the next for a given subject.

**Design**

Based on the TCE, for an empathetic computer system to be robust and accurate, we believe it must have the following properties:

1. It must use a subject-dependent (rather than subject-independent) method of analysis.
2. It must rely heavily on contextual information about the circumstances in which a subject's internal experience is predicted.

In our preliminary implementation we have engineered a system that takes as input a sequence of still facial images (effectively a video) of human subjects and uses information about very specific moments in the input to make predictions about the subject's behavior in the rest of the video.

First, the image sequence is processed by the OpenFace open source software package [3] to track the movement of various key points on a subject's face over time [2,15]. OpenFace then identifies groups of key points moving together in specific way as a facial "Action Unit" (AU) according to the Facial Action Coding System (FACS) [5,7], and analyzes these AUs for their "intensity" (a five-point scale measuring minimum to maximum levels of occurrence) over time [1]. In this way, OpenFace catalogues all movements of the subject's face throughout the video.

Our system then analyzes the AU intensity data to determine when AUs begin, peak (reach an apex of intensity), and end over time, as well as how quickly their intensity rises and falls. Next, the system uses an Affinity Propagation unsupervised clustering algorithm [13] to identify AUs that start and peak close together in time, and groups these co-occurring AUs into an unlabeled "Facial Gesture" (FG). At this point, it is possible to use a simple unsupervised clustering algorithm (such as k-means clustering) to compute which of these FGs are most similar to each other, but we don't just care which FGs are similar to each other – we want to know what those FGs "mean" (or what it is that the subject is expressing via this FG).

Here we introduce contextual information in our model. Our system analyzed videos from the MMSE-HR database of spontaneous expressions [16]. In one half of the database, subjects are performing in a task ("T10") in which a series of three darts are thrown nearer and nearer past/above their head by the experimental facilitator (to provoke a "fear response") [16]. The moments in time when the darts are thrown are not included as part of the dataset but were inferred from visual inspection of subject responses. This simple contextual information allows us to make basic *provable* predictions about a subject's internal experience, based on their facial expressions.

Research efforts on data such as this [10,11,16] often try to "detect" what emotions subjects were feeling during these videos (are these "Fear" responses? "Nervous" responses? "Fun" responses?). The TCE tells us that while subjects were undoubtedly feeling emotions, establishing the "ground truth" of these emotions will be impossible [4]. While expert consensus of trained FACS coders and self-report from subjects might be insightful information, it cannot be *proven to be correct*. There is one thing that can be proven from the videos of the MMSE-HR T10 protocol: when were darts being thrown?

| Early Results | SD | SI |
|---|---|---|
| Top 2 | 37% | 26% |
| Top 10 | 53% | 42% |
| Top 15% | 53% | 32% |
| Median Rank | 4 | 14.5 |

Table 1: Percent of the time our system ranked the correct FG (FGΓ) in the Top 2, Top 10, or Top 15% of all same-subject FGs. The SD method generally outperformed the SI method.

FG rankings are from a median of 48 possible FGs per subject to choose from, among 19 subjects who had significant facial response to protocol T10. Common reasons our system failed to choose FGΓ correctly were:

- FGΓ was "more intense" than FGA and FGβ (the third dart was thrown more closely to subjects' heads).
- Head movement by subject (not yet accounted for by our system).
- Subject made similar FGs in anticipation of a dart throw (a "flinch", or false positive).

### Analysis

We configured our system to analyze the way a subject responds to the first dart and the second dart ((FGA and FGβ, or "FG Alpha and FG Beta"), and to use that (small) training set to try and predict which FG is made in response to the third dart (FGΓ, or "FG Gamma"). To compute FGΓ, our system takes arithmetic means of the multidimensional data from FGA and FGβ, compares all other same-subject FGs to those means, and ranks them from most to least similar using an unsupervised ranking algorithm. Even this basic analysis yields promising results (see Early Results sidebar).

We are unaware of any other research to identify human subject FGs made in response to darts being thrown (with this dataset or others) to which we can directly compare our results, so we also ran a subject-independent (SI) version of our system, in which we compared each subject's FGs to a set of all subjects' FGAs and FGβs, to compare it to our subject-dependent (SD) approach. The SI system was less accurate on average. Our current results support the TCE's claim that no universal or general emotional expression made by humans exists [4], even for such a specific stimulus as this. We believe our current system's accuracy will be further improved by the pending incorporation of subject heart rate and head movement data.

### Conclusion

Since emotions are mental concepts, trying to predict which instance of an emotion a subject is constructing in a given moment is as impossible as trying to "read their thoughts" [4]. While our system *cannot* conclude that a given FG is a subject's "Fear Face", it *can* say that it is likely to be "the face this subject makes when a dart is thrown past their head." That is, by observing FGA and FGβ our system can often predict FGΓ correctly. It could also predict when future darts are being thrown based on the FGs a subject makes, because it knows how they have responded in the past.

Individual subjects' emotive responses to specific stimuli in their environment are not without consistency and can be interpreted correctly given sufficient contextual information, but we must be careful about the conclusions that we draw from these responses. Subject behavior is still meaningful, and we believe our simple method of analysis can be usefully extended into other domains where we have a lot of information about the situation in which subject behavior occurs.

### Future Work

The version of the card game Poker known as Texas Hold 'Em is one such situation that is rich in easy-to-analyze contextual data – the strength of a player's cards, how many chips they have, how much they are betting, etc. – but where the unpredictability of the humans playing remains one of its most exciting and essential elements. The game is almost as much about "playing the player" as it is about "playing the cards".

Our next system will be designed to analyze player behavior in this context. It will provide additional predictive data to one player about their opponent (for instance, are they bluffing?). By learning an opponent's behavior (for example, to discover their "tell" when bluffing) in various "game situations", the system will predict what "game situation" the opponent believes they are in, and therefore the cards they might be holding. If our system can accurately predict the thoughts and feelings of one player, it should give the other a measurable competitive advantage.

**References**

1. Tadas Baltrušaitis, Marwa Mahmoud, Peter Robinson. 2015. Cross-dataset learning and person-specific normalisation for automatic Action Unit detection. In *Proc. 11th IEEE International Conference and Workshops on Automatic Face and Gesture Recognition* (FG '15), 1-6. DOI: 10.1109/FG.2015.7284869.

2. Tadas Baltrušaitis, Peter Robinson, Louis-Philippe Morency. 2013. Constrained Local Neural Fields for Robust Facial Landmark Detection in the Wild. In *Proc. International Conference on Computer Vision Workshops, 300 Faces in-the-Wild Challenge* (ICCVW '13). DOI: 10.1109/ICCVW.2013.54.

3. Tadas Baltrušaitis, Amir Zadeh, Yao Chong Lim, Louis-Philippe Morency. 2018. OpenFace 2.0: Facial Behavior Analysis Toolkit. In *Proc. IEEE International Conference on Automatic Face and Gesture* Recognition (FG '18), 59-66. DOI: 10.1109/FG.2018.00019.

4. Lisa Feldman Barrett. 2017. *How Emotions are Made: The Secret Life of the Brain*. Houghton Mifflin Harcourt Publishing Company, Boston, MA.

5. Paul Ekman, Wallace Friesen, Joseph C. Hager. 2002. *Facial Action Coding System: The Manual on CD ROM.* A Human Face, Salt Lake City, UT.

6. Paul Ekman, Wallace V. Friesen. 1971. Constants Across Cultures in the Face and Emotion. *Journal of Personality and Social Psychology*. 17, 2: 129-149.

7. Carl-Herman Hjortsjö. 1970. *Man's face and mimic language.* Student Research. Universitäts und Landesbibliothek Tirol, Lund, Sweden.

8. Michael Xuelin Huang, Grace Ngai, Kien A. Hua, Stephen Chan, and Hong Va Leong. 2015. Identifying User-Specific Facial Affects from Spontaneous Expressions with Minimal Annotation. *IEEE Transactions on Affective Computing*. 7, 4: 99.

9. G. Littlewort, M. S. Bartlett, I. Fasel, J. Susskind, and J. Movellan. 2004. Dynamics of facial expression extracted automatically from video. In *Proc. Conf. Comput. Vis. Pattern Recog. Workshop* (CVPR '04). 80.

10. Daniel McDuff, Marwa Mahmoud, Ntombikayise Banda, Peter Robinson, Rana el Kaliouby, Rosalind Picard, Tadas Baltrušaitis. 2011. Real-time inference of mental states from facial expressions and body gestures. In *Proc. IEEE Face and Gesture Conference* (FG '11), 1-6. https://doi.org/10.1109/FG.2011.5771372

11. Oveneke, Meshia & Gonzalez, Isabel & Enescu, Valentin & Jiang, Dongmei & Sahli, Hichem. 2017. Leveraging the Bayesian Filtering Paradigm for Vision-Based Facial Affective State Estimation. *IEEE Transactions on Affective Computing,* PP: 1-1.

12. P. Michel and R. El Kaliouby. 2003. Real time facial expression recognition in video using support vector machines. In *Proc. 5th Int. Conf. Multimodal Interfaces* (ICMI '03). 258–264.

13. Pedregosa et al. 2011 Scikit-learn: Machine Learning in Python. *Journal of Machine Learning Research* 12: 2825-2830.

14. M. F. Valstar, M. Mehu, B. Jiang, M. Pantic, and K. Scherer. 2012. Meta- analysis of the first facial expression recognition challenge. In *IEEE Trans. Syst. Man. Cybern. B. Cybern.* 42, 4: 966–979.

15. Amir Zadeh, Tadas Baltrušaitis, Louis-Philippe Morency. 2017. Convolutional Experts Constrained Local Model for Facial Landmark Detection. In *Proc. IEEE Conference on Computer Vision and Pattern Recognition Workshops* (CVPRW '17), 2051-2059. DOI: 10.1109/CVPRW.2017.256.

16. Zheng Zhang, Jeff Girard, Yue Wu, Xing Zhang, Peng Liu, Umur Ciftci, Shaun Canavan, Michael Reale, Andy Horowitz, Huiyuan Yang, Jeff Cohn, Qiang Ji, and Lijun Yin. 2016. *Multimodal Spontaneous Emotion Corpus for Human Behavior*

*Analysis*. In *Proc. IEEE International Conference on Computer Vision and Pattern Recognition* (CVPR '16). DOI: 10.1109/CVPR.2016.374.

17. Mingmin Zhao, Fadel Adib, and Dina Katabi. 2016. Emotion recognition using wireless signals. In *Proc. 22nd Annual International Conference on Mobile Computing and Networking* (MobiCom '16). ACM, New York, NY, USA, 95-108. DOI: https://doi.org/10.1145/2973750.2973762